\documentclass[runningheads]{llncs}

\usepackage{epsfig}
\usepackage{graphicx}
\usepackage{comment}
\usepackage{amsmath,amssymb,amsfonts} 

\usepackage{color, soul}
\usepackage{enumerate}
\usepackage[shortlabels]{enumitem}
\usepackage[dvipsnames]{xcolor}
\usepackage[nocompress]{cite} 
\usepackage{tabu}
\usepackage{xpunctuate}
\usepackage{comment}
\usepackage{multirow}
\usepackage{booktabs}
\usepackage{pifont}
\usepackage[pagebackref=true,breaklinks=true,letterpaper=true,colorlinks,bookmarks=false]{hyperref}
\usepackage{algorithm}
\DeclareFixedFont{\ttb}{T1}{txtt}{bx}{n}{6} 
\DeclareFixedFont{\ttm}{T1}{txtt}{m}{n}{6}  

\usepackage{color}
\definecolor{deepblue}{rgb}{0,0,0.5}
\definecolor{deepred}{rgb}{0.6,0,0}
\definecolor{deepgreen}{rgb}{0,0.5,0}

\usepackage{xcolor}

\definecolor{codegreen}{rgb}{0,0.6,0}
\definecolor{codegray}{rgb}{0.5,0.5,0.5}
\definecolor{codepurple}{rgb}{0.58,0,0.82}
\definecolor{backcolour}{rgb}{0.95,0.95,0.92}

\usepackage{listings}

\definecolor{codeblue}{rgb}{0.25,0.5,0.5}
\DeclareMathAlphabet{\mathpzc}{OT1}{pzc}{m}{it}

\providecommand{\MDPC}[0]{{\tt MemDPC}}
\providecommand{\ie}[0]{\emph{i.e\xperiod}}
\providecommand{\eg}[0]{\emph{e.g\xperiod}}
\providecommand{\etal}[0]{\emph{et al\xperiod}}

\usepackage[dvipsnames]{xcolor}
\definecolor{mygray}{gray}{0.4}

\usepackage{pifont}
\newcommand{\cmark}{\ding{51}}%
\newcommand{\xmark}{\ding{55}}%

\usepackage[width=122mm,left=12mm,paperwidth=146mm,height=193mm,top=12mm,paperheight=217mm]{geometry}

\begin{document}
\pagestyle{headings}
\mainmatter
\def\ECCVSubNumber{366}  

\title{Memory-augmented Dense Predictive Coding for Video Representation Learning} 

\titlerunning{Memory-augmented Dense Predictive Coding}
%
\author{Tengda Han \and
Weidi Xie  \and
Andrew Zisserman }
\authorrunning{T. Han et al.}
%
\institute{
Visual Geometry Group, Department of Engineering Science, University of Oxford\\
\email{\{htd,weidi,az\}@robots.ox.ac.uk}}
\maketitle

\begin{abstract}
The objective of this paper is self-supervised learning from video, in particular for representations for action recognition.
We make the following contributions:
(i) We propose a new architecture and learning framework 
{\em Memory-augmented Dense Predictive Coding} (\MDPC) for the task.
It is trained with a \emph{predictive attention mechanism} over the set of \emph{compressed memories}, 
such that any future states can always be constructed by a convex combination of the condensed representations,
allowing to make multiple hypotheses efficiently.
(ii) We investigate visual-only self-supervised video representation learning from RGB frames, 
or from unsupervised optical flow, or both.
(iii) We thoroughly evaluate the quality of the learnt representation on four different downstream tasks:
~action recognition, video retrieval, learning with scarce annotations, and unintentional action classification.
In all cases, 
we demonstrate state-of-the-art or comparable performance over other approaches with orders of magnitude fewer training data.
\end{abstract}

\stepcounter{footnote}
\footnotetext{Code is available at~\url{http://www.robots.ox.ac.uk/\textasciitilde vgg/research/DPC}}
\section{Introduction}
Recent advances in self-supervised representation learning for images have yielded impressive 
results, \eg~\cite{Oord18,Henaff19,Hjelm19,Tian19,Ji19,He20,Zhuang19,Chen20}, 
with performance matching or exceeding that of supervised representation learning on downstream tasks.
However, in the case of videos, although there have been similar gains for 
{\em multi-modal} self-supervised representation learning, \eg~\cite{Arandjelovic17,Korbar18,Sun19,Alwassel19,Piergiovanni20,Miech20}, 
progress on learning {\em only} from the video stream (without additional audio or text streams) 
is lagging behind. The objective of this paper is to improve the performance of video only self-supervised learning.

Compared to still images, 
videos should be a more suitable source for self-supervised representation learning
as they naturally provide various augmentation, such as object out of plane rotations and deformations.
In addition, 
videos contain additional temporal information that can be used to disambiguate actions
\eg~open vs.~close.
The temporal information can also act as a free supervisory signal to train a model
to predict the future states from the past either passively by watching videos~\cite{Vondrick16b,Lotter17,Han19}
or actively in an interactive environment~\cite{Dosovitskiy17}, and thereby learn a video representation.

\begin{figure}[h]
	\centering
	\includegraphics[width=\textwidth]{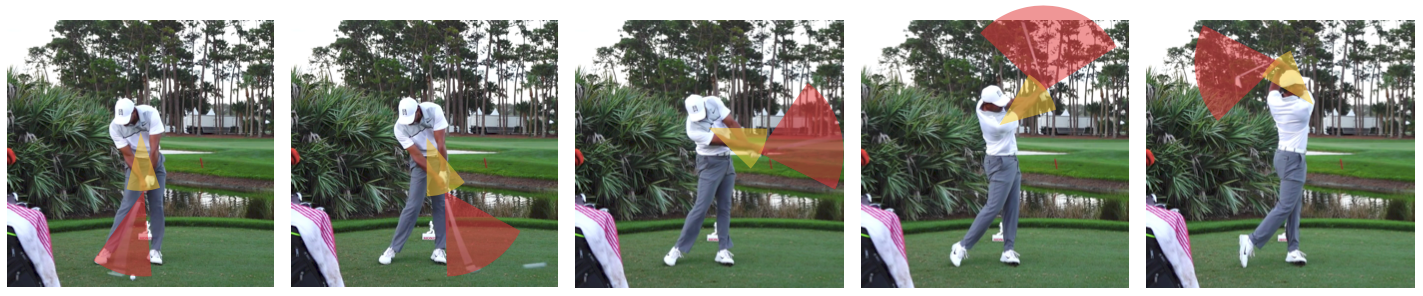}
	\caption{\textbf{Can you predict the next frame?} 
	Future prediction naturally involves challenges from multiple hypotheses, 
	\eg~the motion of each leaf, reflections on the water, hands and the golf club can be in many possible positions}
	\label{fig:golf}
\end{figure}

Unfortunately, the exact future is indeterministic (a problem long discussed in the history of science, and
known  as ``Laplace's Demon''). 
As shown in Figure~\ref{fig:golf}, 
this problem is directly apparent in the stochastic variability of scenes, 
\eg~trying to predict the exact motion of each leaf on a tree when the wind blows, or the changing reflections on the water.
More concretely, consider the action of  `playing golf' -- once the action starts, 
a future frame could have the hands and golf club in many possible positions,
depending on the person who is playing.
Learning visual representation by predicting the future therefore requires designing specific training schemes that 
simultaneously circumvents the unpredictable details in exact frames, 
and also handles multiple hypotheses and incomplete data -- 
in particular only one possible future is exposed by the frames of one video.

Various approaches have been developed to deal with the multiple possible futures for an action.
Vondrick~\etal~\cite{Vondrick16b}  explicitly generates multiple hypotheses, 
and only the hypothesis that is closest to the true observation is chosen during optimization,
however, this approach limits the number of possible future states.
Another line of work~\cite{Oord18,Han19} circumvents this difficulty by using contrastive learning --
the model is only asked to predict \emph{one} future state that 
assigns higher similarity to the true observation than to any distractor observation 
(from different videos or from elsewhere in the same video).
Recalling the `playing golf' example, 
the embedding must capture the hand movement for this action, 
but not necessarily the precise position and velocity, only sufficiently to disambiguate future frames.

In this paper, we continue the idea of contrastive learning,
but improve it by the addition of a {\em Compressive Memory},
which maps ``lifelong'' experience to a set of compressed memories and helps to better anticipate the future.
We make the following four contributions:
\emph{First}, 
we propose a novel Memory-augmented Dense Predictive Coding (\MDPC) architecture.
It is trained with a \emph{predictive attention mechanism} over the set of \emph{compressed memories}, 
such that any future states can always be constructed by a convex combination of the condensed representations,
allowing it to make multiple hypotheses efficiently.
\emph{Second}, we investigate visual only self-supervised video representation learning from RGB frames, 
or from unsupervised optical flow, or both.
\emph{Third}, we argue that, 
in addition to the standard linear probes and fine-tuning~\cite{Zhang16color,Sun19},
that have been used for evaluating representation from self-supervised learning,
a non-linear probe should also be used, 
and demonstrate the difference that this probe makes.
\emph{Finally},
we evaluate the quality of learnt feature representation on four different downstream tasks:
action recognition, learning under low-data regime~(scarce annotations), video retrieval, and unintentional action classification;
and demonstrate state-of-the-art performance over other approaches with similar settings on \emph{all} tasks.

\section{Related Work}

\noindent {\bf Self-supervised learning for images}
has undergone rapid progress  in visual representation learning recently~\cite{Oord18,Henaff19,Hjelm19,Tian19,Ji19,He20,Zhuang19,Chen20}.
Generally speaking, the success can be attributed to one specific training paradigm,
namely contrastive learning~\cite{Chopra05,Gutmann10}, \ie~contrast the positive and negative sample pairs.\\

\noindent {\bf Self-supervised learning for videos} has explored various ideas 
to learn representations by exploiting spatio-temporal
information~\cite{Isola15,Agrawal15,Jayaraman15,Wang15,Misra16,Vondrick16b,Jia16,Jayaraman16,Lotter17,Fernando17,Lee17,Wiles18a,Gan18,Jakab18,Jing18,Vondrick18,Kim19,Xu19vcop,Diba19,Lai19,Wang19,Alwassel19,Lai20}.  Of more relevance here
is the line of research using contrastive learning, 
\eg~\cite{Arandjelovic17,Arandjelovic18,Korbar18,Alwassel19,Patrick20,Piergiovanni20} learn from visual-audio correspondence, 
\cite{Miech20} learns from video and narrations,
and our previous work~\cite{Han19} learns video representations by predicting future states.\\

\noindent {\bf Memory models}
have been considered as one of the fundamental building blocks towards intelligence.
In the deep learning literature,
two different lines of research have received extensive attention,
one is to build networks that involve an internal memory which can be implicitly updated in a recurrent manner, 
\eg~LSTM~\cite{Hochreiter97} and GRU~\cite{Chung14J}.
The other line of research focuses on augmenting feed-forward models with an explicit memory that can be read or written to with an attention-based procedure~\cite{Xu15,Vaswani17,Bahdanau15,Wang18,Devlin18,Kumar16,Graves14a,Sukhbaatar15}. 
In this work,
our compressive memory falls in the latter line, \ie{}~an external memory module.
\section{Methodology}
\label{sec:method}
The proposed Memory-augmented Dense Predictive Coding~(\MDPC), 
is a conceptually simple model for learning a video representation with contrastive predictive coding. 
The key novelty is to
augment the previous DPC model with a \emph{Compressive Memory}.
This provides a mechanism for handling the multiple future hypotheses required in learning due to
the problem that only one possible future is exposed by a particular video.

\begin{figure}[h]
	\centering
	\includegraphics[width=\textwidth]{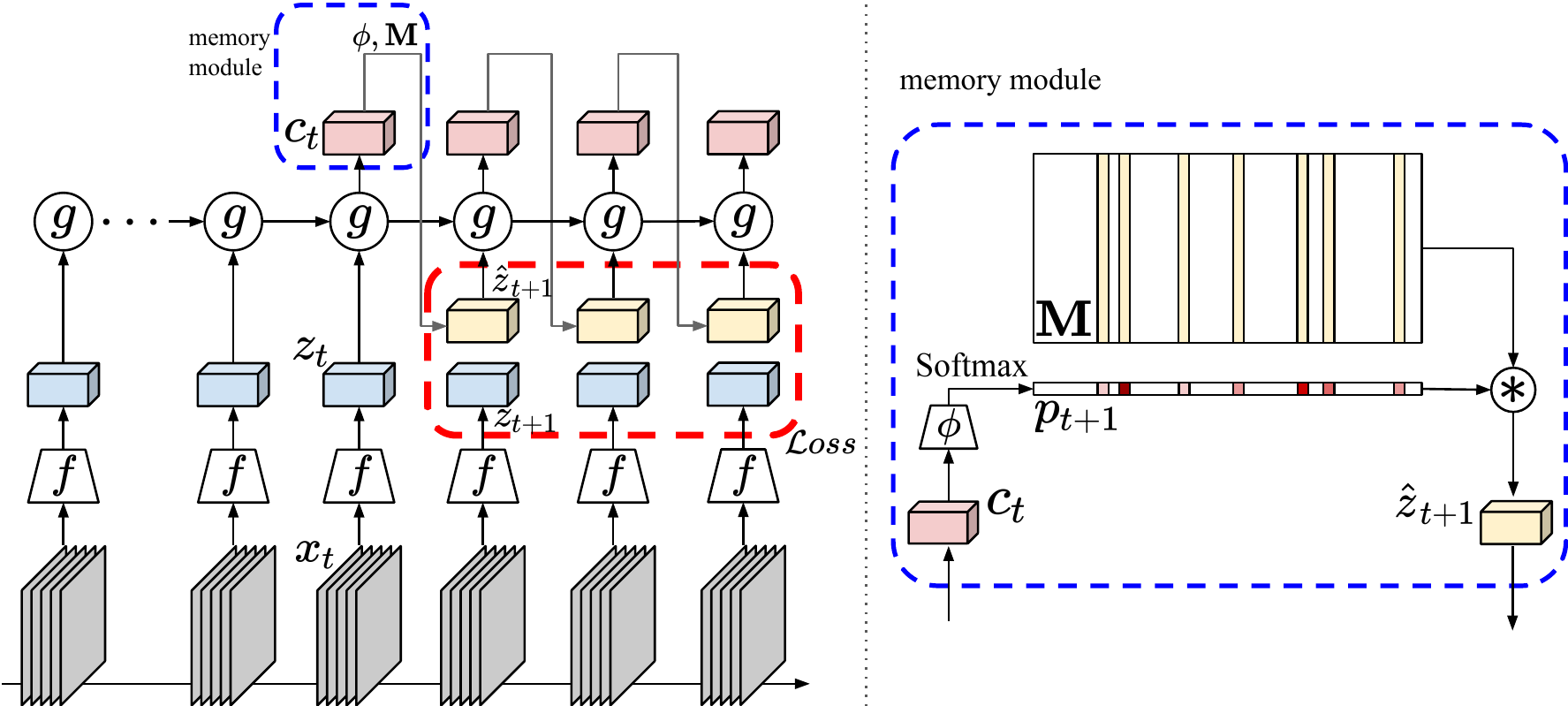}
	\caption{Architecture of the Memory-augmented Dense Predictive Coding (\MDPC). Note, the memory module is only used during the self-supervised training. The $c_t$ embedding is used for downstream tasks}\label{fig:arch-mem}
	\label{fig:mdpc}
\end{figure}

The architecture is shown in Figure~\ref{fig:mdpc}. 
As in the case of DPC, 
the video is partitioned into 8 blocks with 5 frames each, 
and an encoder network $f$ generates an embedding $z$ for each block. 
For inference, these embeddings are aggregated over time by a function $g$ into a video level embedding $c$. 
During training, 
the future block embeddings $\hat{z}$ are predicted and used to select the true embedding in the dense predictive coding manner. 
In \MDPC{}, the prediction of $\hat{z}$ is from a convex combination
of memory elements (rather from $c$ directly as in DPC), 
and it is this restriction that also enables the network to handle multiple hypotheses, as will be explained below.

\subsection{Memory-augmented Dense Predictive Coding~(\MDPC)}
\label{sec:mdpc}
\par{\bf Video Block Encoder.}
As shown in Figure~\ref{fig:arch-mem}, 
we partition the input video sequence into multiple blocks $x_1, ..., x_t, x_{t+1}, ...$, 
where each block is composed of multiple video frames. 
Then a shared feature extractor $f(.)$~(architecture details are given in the appendix) 
extracts the video features $z_{i}$ from each video block $x_{i}$:
\begin{equation}
z_i = f(x_i)
\end{equation}

\noindent {\bf  Temporal Aggregation.}
After acquiring block  representations,
multiple block embeddings are aggregated to obtain a context feature $c_t$,  
summarizing the information over a longer temporal window. 
Specifically, 
\begin{equation}
c_t = g(z_1, z_2, ..., z_t)
\end{equation}
in our case, we simply adopt Recurrent Neural Networks (RNNs) for $g(.)$,
but other auto-regressive model should also be feasible for temporal aggregation.\\

\noindent {\bf Compressive Memory.}
In order to enable efficient multi-hypotheses estimation,
we augment the predictive models with an external common compressive memory.
This external memory bank is shared for the entire dataset during training,
and is accessed by a \emph{predictive addressing mechanism} 
that infers a probability distribution over the memory entries, 
where each memory entry acts as a potential hypothesis.

In detail, 
the compressed memory bank is written
$\mathbf{M} = [m_1,m_2,...,m_k]^\top \in\mathbb{R}^{k\times C}$,
where $k$ is the memory size and $C$ is the dimension of each compressed memory. 
During training,
a predictive memory addressing mechanism~(Eq.~\ref{eq:pmam}) is used to draw a hypothesis from the compressed memory, 
and the predicted future states~$\hat{z}_{t+1}$  is then computed as the expectation of sampled hypotheses~(Eq.~\ref{eq:esh}):
\begin{align}
p_{t+1} &= \text{Softmax}\big(\phi(c_t)\big)  \label{eq:pmam} \\
\hat{z}_{t+1} &= \sum_{i=1}^{k} p_{(i, t+1)} \cdot m_i = p_{t+1} \mathbf{M} \label{eq:esh} 
\end{align}
\noindent where $p_{(i,t+1)} \in \mathbb{R}^{k}$ refers to the contribution of $i$-th memory slot for the future representation at time step $t$. 
A future prediction function $\phi(.)$ projects the context representation to $p_{(i,t+1)}$,
in practice, $\phi(.)$ is learned with a multilayer perceptron. 
The softmax operation is applied on the $k$ dimension. \\

\noindent {\bf  Multiple Hypotheses.}
The dot product of the predicted and desired future pairs can be rewritten as:
\begin{equation}
\label{eq:dot}
\psi(\hat{z}^{\top}, z) = \bigg(\sum_{i=1}^{k}{p_i \cdot m_i^{\top}}\bigg) z = \sum_{i=1}^{k}{p_i \cdot \big(m_i^{\top} z\big)}
\end{equation}
\noindent where $m_i^{\top} z$ refers to the dot product~(\ie~similarity) between a single memory slot 
and the feature states from the observation. 
The objective of $\phi(.)$ is to predict a probability distribution over $k$ hypotheses in the memory bank, 
such that the expectation of $m_i^{\top} z$ for a positive pair is larger than that of negative pairs.
Since the future is uncertain, 
the desired future feature $z$ might be similar to one of the multiple hypotheses in the memory bank, 
say either $m_p$ or $m_q$, for instance. 
To handle this uncertainty, 
the future prediction function $\phi(.)$ just needs  to put a higher probability on both the  $p$ and $q$ slots, 
such that Equation~\ref{eq:dot} is always large no matter which state the future is. 
In this way, 
the burden of modelling the future uncertainly is allocated to the memory bank $\mathbf{M}$ and future prediction function $\phi(.)$, 
thus the backbone encoder $f(.)$ and $g(.)$ can save capacity and capture the high-level action trajectory.\\

\noindent {\bf  Memory Mechanism Discussion.}
Note, in contrast to the memory mechanism in Wu~\etal{}~\cite{Wu18} and MoCo~\cite{He20}, 
which has the goal of storing more data samples to increase the number of negative samples during contrastive learning,
our Compressive Memory has the goal of aiding learning by compressing 
all the potential hypotheses within the entire dataset, and allowing 
access through the  predictive addressing mechanism.
The  memory mechanism shares similarity with NetVLAD~\cite{Arandjelovic16},
which represents a feature distribution with ``trainable centroids''. However, 
in NetVLAD the goal is for compact and discriminative feature aggregation, and it encodes
a weighted sum of {\em residuals}  between feature vectors and the centroids.
In contrast, our goal with  $\phi(.)$ is to predict attention weights for the entries in the memory bank $\mathbf{M}$,
in order to construct the  future state  as a convex combination these entries. 
The model can also sequentially predict further into the future with the \emph{same} memory bank.

\subsection{Contrastive Learning}
\label{sec:contrast}
Contrastive Learning generally refers to the training paradigm that forces 
the similarity scores of positive pairs to be higher than those of negatives. 
\begin{figure}[h]
	\centering
	\includegraphics[width=0.7\textwidth]{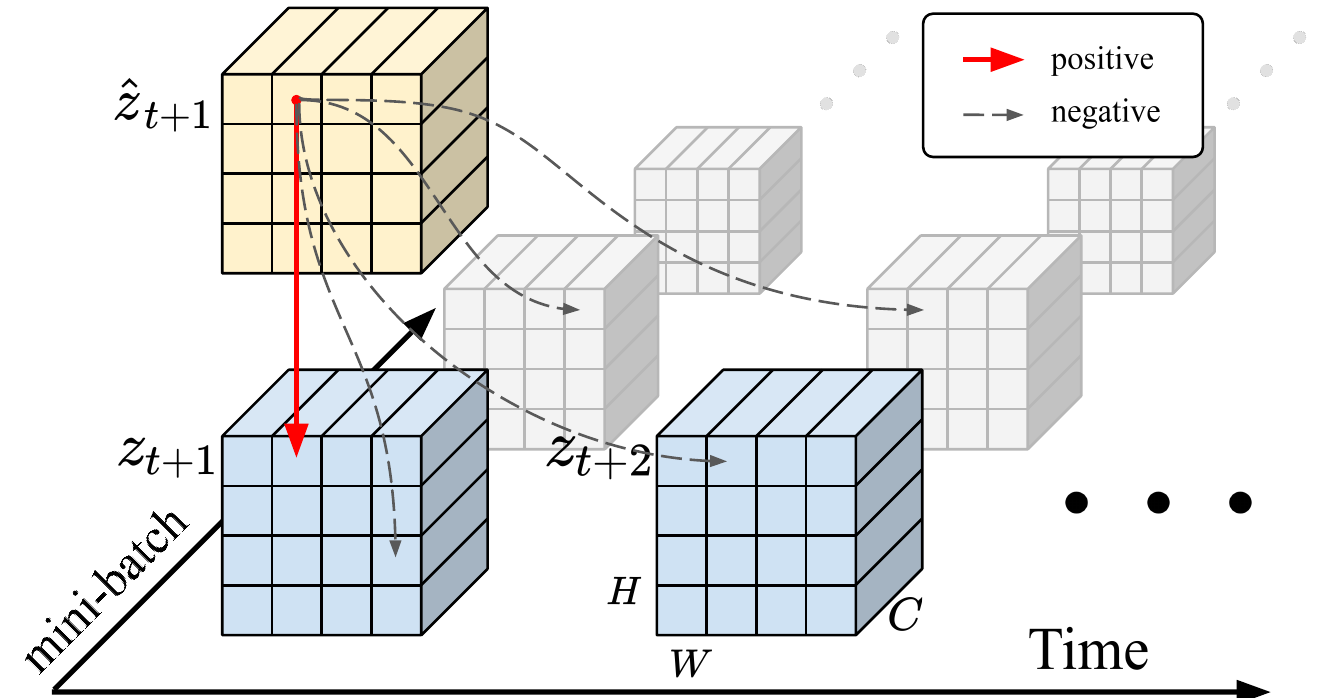}
	\caption{Details of the contrastive loss. Contrastive loss is computed densely, \ie{} over both spatial and temporal dimension of the feature}\label{fig:contrastive}
\end{figure}

Specifically, in \MDPC{},
we predict the future states recursively, 
ending up with a sequence of predicted features $\hat{z}_{t+1}, \hat{z}_{t+2}, \dots, \hat{z}_{end}$ 
and the video feature from the true observations~$z_{t+1}, z_{t+2}, \dots, z_{end}$.
As shown in Figure~\ref{fig:contrastive},
each predicted $\hat{z}$ in practise is a dense feature map.
To simplify the notation, 
we denote temporal index with $i$ and denote other indexes including spatial index and batch-wise index as $k$, 
where batch-wise index means the index in the current mini-batch, $k\in \{(1,1,1), (1,1,2), ..., (B,H,W)\}$. 
The objective function to minimize becomes: 
\begin{align}
\mathcal{L} &= -\mathbb{E} \left[ \sum_{i,k} \log 
\frac{ e^ {\psi ( \hat{z}_{i,k}^\top, z_{i,k})}}
{e^{\psi(\hat{z}_{i,k}^\top, z_{i,k})} + \sum_{(j,m)\neq(i,k)} e^{\psi( \hat{z}_{i,k}^\top, z_{j,m})}} \right] 
\label{eq:infonce}
\end{align}
where $\psi(\cdot)$ is acting as a \emph{critic} function, in our case, we simply use dot product between the two vectors~(we also experiment with L2-normalization,
and find it gives similar performance on downstream tasks).
Essentially, the objective function acts as a multi-way classifier, 
and the goal of optimization is to learn the video block encoder that assigns the highest values for $(\hat{z}_{i,k},z_{i,k})$ 
\ie~higher similarity between the predicted future states and that from true observations originating from the same video and spatial-temporal aligned position. 

\subsection{Improving  Performance by Extensions}
\label{sec:bd_flow}
As  \MDPC{} is a general self-supervised learning framework, it can be combined with other `modules'  like two-stream networks and bi-directional RNN to improve the quality of the visual representations. \\

\noindent {\bf Two-stream Architecture.} 
We represent dense optical flow as images by stacking the $x$ and $y$ displacement fields and another zero-valued array to make them 3-channel images. There is no need to modify the \MDPC{} framework, 
and it can be directly applied to optical flow inputs by simply replacing the input $x_t$ from RGB  frames to optical flow frames. 
We use late fusion like~\cite{Simonyan14b,Feichtenhofer16} to combine both streams.\\

\noindent {\bf  Bi-directional \MDPC.}
From the perspective of human perception, where only the future is actively predicted, 
\MDPC{} is initially designed to be single-directional. However, when passively taking the videos as input, 
predicting backwards becomes feasible. Bi-directional \MDPC{} has a shared feature extractor $f(.)$ to extract the features $z_{1}, z_{2}, ... , z_{t}$, but has two identical aggregators $g^{f}(.)$ and $g^{b}(.)$ denoting forward and backward aggregation. They aggregate the bi-directional context features $c^{f}_{t}$ and $c^{b}_{t}$. 
Then \MDPC{} predicts the past and the future features with the shared $\phi(.)$ and shared memory bank $\mathbf{M}$, 
and constructs contrastive losses for both directions, namely $\mathcal{L}^{f}$ and $\mathcal{L}^{b}$. The final loss is the average of the losses from both directions.  

\section{How to Evaluate Self-Supervised Learning?}
The standard way to evaluate the quality of the learned representation is to assess the performance on
downstream tasks using two protocols: 
(i) a linear probe -- freezing the network and only train a linear head for the downstream task;
or (ii) fine-tuning the entire network for the downstream task. 
For example, in (i) if the downstream task is classification, \eg{} of UCF101, 
then a linear classifier is trained on top of the frozen base network. 
In (ii) the self-supervised training of the base network only provides the initialization.
However,  there is no particular reason why self-supervision should lead to features that are linearly separable, 
even if the representation has encoded semantic information.
Consequently, in addition to the two protocols mentioned above, 
we also evaluate the frozen features with non-linear probing,
\eg~in the case of a classification downstream task, a non-linear MLP head is trained as the final classifier.
In the experiments we evaluate the representation on four different downstream tasks.\\

\begin{figure}[h]
    \centering
    \includegraphics[width=\textwidth]{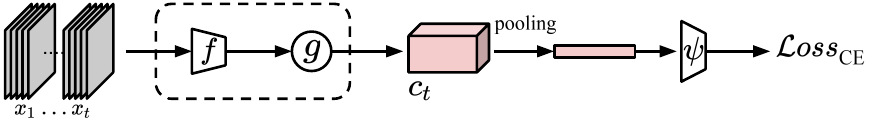}
    \caption{Architecture of the action classification framework}\label{fig:classification-arch}
\end{figure}

\noindent {\bf   Action Classification}
is a common evaluation task for self-supervised learning on videos and it allows us to compare against other methods. 
After self-supervised training,
our \MDPC{} can be evaluated on this task under two settings:
(i) linear and non-linear probing with a fixed network (here the entire backbone network, namely $f(.)$, $g(.)$);
and  (ii) fine-tuning the entire network end-to-end with supervised learning.
For the embedding, as shown in Figure~\ref{fig:classification-arch}, 
we take the input video blocks $x_1, x_2, ..., x_t$ in the same way as \MDPC\ 
and extract the context feature $c_t$ using the  feature extractor $f(.)$ for each block and temporal aggregator $g(.)$;
then we spatially pool the context feature $c_t$ to obtain the embedding.
We describe the training details in Section~\ref{exp:sup_detail}.
The detailed experiment can be found in Section~\ref{exp:action}. \\

\noindent {\bf  Data Efficiency and Generalizability} 
are reflected by the effectiveness of the representation under a scarce-annotation regime.
For this task, 
we take the \MDPC\ representation and finetune it for action classification task,
but limit the model to only use 10\%, 20\% and 50\% of the labelled training samples, 
then we report the accuracy on the same testing set. 
The classifier has the identical training pipeline as shown in Figure~\ref{fig:classification-arch}, 
and training details are explained in Section~\ref{exp:sup_detail}. 
The detailed experiment can be found in Section~\ref{exp:efficiency}.\\

\noindent {\bf   Video Action Retrieval}
directly evaluates the quality of the representation without any further training,
aiming to provide a straightforward understanding on the quality of the learnt representation.
Here, we use the simplest non-parametric classifier, 
\ie~k-nearest neighbours, to determine whether semantically similar actions are close in the high-dimensional space.
Referring to Figure~\ref{fig:classification-arch},
for each video, 
we truncate it into blocks $x_1, x_2, ..., x_t$ and extract the context feature $c_t$ with the $f(.)$ and $g(.)$ trained with \MDPC. 
We spatially pool $c_t$ to get a context feature vector, 
which is directly used as a query vector for measuring the similarity with other videos in the dataset. 
The detailed experiment can be found in Section~\ref{exp:retrival}.\\

\noindent {\bf  Unintentional Actions}
is a straightforward application for a predictive framework like \MDPC.
We evaluate our representation on the task of unintentional event classification that is proposed in the recent Oops dataset~\cite{Epstein20}.
The core of unintentional events detection in video is a problem of anomaly detection.
Usually, one of the predicted hypotheses tends to match true future relatively well for most of the videos. 
The discrepancy between them yields a measurement of future predictability, or `surprise' level.
A big surprise or a mismatched prediction can be used to locate the failing moment.
In detail, we design the model as follows: 
first, we compute both the predicted feature $\hat{z}_{i}$ and the corresponding true feature $z_{i}$, 
and let a function $\xi(.)$ to measure their discrepancy. 
We train the model with two settings: 
(i) freezing the representation and only train the classifier $\xi(.)$; 
(ii) finetuning the entire network. 
The detailed structure for the classification task can be found in Section~\ref{sec:oops}.

\section{Experiments}
\label{sec:exp}

\subsection{Datasets}
For the self-supervised training, 
two video action recognition datasets are used, but labels are dropped during training:
{\em UCF101}~\cite{Soomro12}, containing 13k videos spanning over 101 human actions;
and Kinetics400 ({\em K400})~\cite{Kay17} with 306k 10-second video clips covering 400 human actions.
For the downstream tasks we also use UCF101, and additionally we use:
{\em HMDB51}~\cite{Kuehne11} containing 7k videos spanning over 51 human actions; 
and~{\em Oops}~\cite{Epstein20} containing 20k videos of daily human activities with unexpected failed moments, 
among them 14k videos have the time stamps of the failed moments manually labelled.

\subsection{Self-Supervised Training}
In our experiment, 
we use a (2+3D)-ResNet, following~\cite{Feichtenhofer18,Han19}, as the encoder $f(.)$, 
where the first two residual blocks res2 and res3 have 2D convolutional kernels, 
and only res4 and res5 have 3D kernels.
Specifically, (2+3D)-ResNet18 and (2+3D)-ResNet34 are used in our experiments, 
denoted as R18 and R34 below. 
For the temporal aggregation, $g(\cdot)$, 
we use an one-layer GRU with kernel size $1\times1$, 
with the weights shared among all spatial positions on the feature map. 
The future prediction function, $\phi(.)$,  is a two-layer convolutional network. 
We choose the size of the  memory bank $\mathbf{M}$ to be 1024 based on experiments in Table~\ref{table:ablation}. 
Network architecture are given in the appendix.

For the data, raw videos are decoded at a frame rate 24-30 fps, 
and each data sample consists of 40 consecutive frames, 
sampled with a temporal stride of 3 from the raw video. 
As input to \MDPC{}, 
they are divided into 8 video clips -- so that each encoder $f(.)$ inputs 5 frames, covering around 0.5 seconds,
and the 40 frames around 4 seconds.
For optical flow, in order to eliminate extra supervisory signals in the self-supervised training stage, 
we use the \textbf{un-supervised} TV-L1 algorithm~\cite{Zach07},
and follow the same pre-processing procedures as~\cite{Carreira17}, 
\ie~truncating large motions with more than $\pm 20$ in both channels, 
appending a third $0$s channel, and transforming  the values  from $[-20,20]$ to $[0,255]$.
For data augmentation, 
we apply clip-wise random crop and horizontal flip, 
and frame-wise color jittering and random greyscale, for both the  RGB and optical flow streams.
We experiment with both $128\times128$ and $224\times224$ input resolution.
The original video resolution is $256\times256$ and it is firstly cropped to $224\times224$ then rescaled if needed.  
Self-supervised training uses the Adam~\cite{Kingma19} optimizer with initial learning rate $10^{-3}$.
The learning rate is decayed once to $10^{-4}$ when the validation loss plateaus. 
We use a batch size of 16 samples per GPU. 

\subsection{Supervised Classification}
\label{exp:sup_detail}

For all action classification downstream tasks,
the input follows the same frame sampling procedure as when the model is trained with self-supervised learning,
and then we train the classifier with cross-entropy loss as shown in Figure~\ref{fig:classification-arch}.
A dropout of 0.9 is applied on the final layer.
For data augmentation, 
we use clip-wise random crop, random horizontal flip, and random color jittering.
The classifier is trained with Adam with a $10^{-3}$ initial learning rate, 
and decayed once to $10^{-4}$ when the validation loss plateaus. 
During testing, we follow the standard pipeline, 
\ie~ten-crop (center and four corner crops, w/o horizontal flip), 
take the same sequence length as training from the video,
and average the prediction from the sampling temporal moving window.

\subsection{Evaluation: \ Action Classification}
\label{exp:action}
We conduct two sets of  experiments:
(i) ablation studies on the effectiveness of the different modules in the \MDPC,
by self-supervised learning on UCF101,
(ii) to compare with other state-of-the-art approaches, 
we run \MDPC{} on K400 with self-supervised learning.
For both settings, 
the representation quality is evaluated on UCF101 and HMDB51 with linear probing, 
non-linear probing,  and end-to-end finetuning.

\subsubsection*{Ablations on UCF101.}
\label{exp:ucf_ablation}
In this section, 
we conduct extensive experiments to validate the effectiveness of compressive memory, 
bidirectional aggregation,  and self-supervised learning on optical flow.
Note that, in each experiment, we keep the settings identical, and only vary one variable at a time.

As shown in Table~\ref{table:ablation}, the following phenomena can be observed:
\emph{First},
comparing experiment $\mathcal{C}2$ against $\mathcal{B}1$~($68.2$ vs.~$61.8$), 
networks initialized with self-supervised \MDPC\ clearly present better generalization than a randomly initialized network;
\emph{Second}, 
comparing with a strong baseline~($\mathcal{A}$),
the proposed compressive memory boost the learnt representation by around $5\%$~($68.2$ vs.~$63.6$),
and the optimal memory size for UCF101 is $1024$;
\emph{Third},
\MDPC{} acts as a general learning framework that can also help to boost the generalizability of motion representations,
a $7.3\%$ boost can be seen from $\mathcal{D}1$ vs.~$\mathcal{B}2$~($81.9$ vs.~$74.6$);
\emph{Fourth},
the bidirectional aggregation provides a small boost to the accuracy by about 1\% ($\mathcal{E}1$ vs. $\mathcal{C}2$, $\mathcal{E}2$ vs. $\mathcal{D}1$, $\mathcal{E}3$ vs. $\mathcal{D}2$).
\emph{Lastly},
after fusing both streams, $\mathcal{D}2$ achieves 84\% classification accuracy,
confirming our claim that self-supervised learning with only the video stream~(without additional audio or text streams) 
can also end up with strong action recognition models.

\begin{table}[!h]
	\scriptsize
	\centering
	\caption{Ablation studies. 
	We train \MDPC\ on UCF101 and evaluate on UCF101 action classification by finetuning the entire network. 
	We group rows for clarity: $\mathcal{A}$ is the reimplementation of DPC, $\mathcal{B}$ are random initialization baselines, 
	$\mathcal{C}$ for different memory size, $\mathcal{D}$ incorporates optical flow, $\mathcal{E}$ incorporates a  bi-directional RNN 
	}
	\label{table:ablation}
	\begin{tabular}{l|lllll|l}
		\hline
		\multirow{2}{*}{\#} & \multirow{2}{*}{Network} & \multicolumn{4}{c|}{Self-Sup.}  & \multicolumn{1}{c}{Sup. (top1)} \\
		   &                     & Dataset         & Input   & Resolution       & Memory size      & UCF101(ft)       \\ \hline
		$\mathcal{A}$  & R18                 & UCF101          & RGB     & $128 \times 128$ & -~(DPC~\cite{Han19})& 63.6             \\  \hline
		
		$\mathcal{B}1$ & R18                 & -(rand. init.)  & RGB     & $128 \times 128$ & -      & 61.8                 \\ 
		$\mathcal{B}2$ & R18                 & -(rand. init.)  & Flow    & $128 \times 128$ & -      & 74.6                 \\
		$\mathcal{B}3$ & R18$\times2$        & -(rand. init.)  & RGB+F   & $128 \times 128$ & -      & 78.7                 \\ \hline
		
		$\mathcal{C}1$ & R18                 & UCF101          & RGB     & $128 \times 128$ & 512    & 65.3                 \\
		$\mathcal{C}2$ & R18                 & UCF101          & RGB     & $128 \times 128$ & 1024   & {68.2}     \\
		$\mathcal{C}3$ & R18                 & UCF101          & RGB     & $128 \times 128$ & 2048   & 68.0          \\ \hline
		$\mathcal{D}1$ & R18                 & UCF101          & Flow    & $128 \times 128$ & 1024   & {81.9}     \\
		$\mathcal{D}2$ & R18$\times2$        & UCF101          & RGB+F   & $128 \times 128$ & 1024   & {84.0}     \\\hline
		
		$\mathcal{E}1$ & R18-bd              & UCF101          & RGB     & $128 \times 128$ & 1024  & {69.2}         \\ 
		$\mathcal{E}2$ & R18-bd              & UCF101          & Flow    & $128 \times 128$ & 1024  & 82.3                  \\ 
		$\mathcal{E}3$ & R18-bd$\times 2$    & UCF101          & RGB+F   & $128 \times 128$ & 1024  & 84.3  \\ \hline
	\end{tabular}
\end{table}

\subsubsection*{Comparison with others.}
\label{sec:ucf_sota}
In this section, 
we train \MDPC\ on K400 and evaluate the action classification performance on UCF101 and HMDB51.
Specifically, we evaluate three settings: 
(1) finetuning the entire network (denoted as Freeze=\xmark);
(2) freeze the backbone and only train a linear classifier, \ie~linear probe~(denoted as Freeze=\cmark);
(3) freeze the backbone and only train a \textbf{non-linear} classifier, \ie~non-linear probe~(denoted as `n.l.'). 

\setlength{\tabcolsep}{1pt}
\begin{table}[!htb]
\scriptsize
	\centering
	\caption{\scriptsize{Comparison with state-of-the-art approaches.
		In the left columns, we show the pre-training setting, \eg~dataset, resolution, architectures with encoder depth, modality.
		In the right columns, the top-1 accuracy is reported on the downstream action classification task for different datasets, 
		\eg~UCF, HMDB, K400.
		The dataset parenthesis shows the total video duration in time (\textbf{d} for day, \textbf{y} for year).
		`Frozen \xmark' means the network is end-to-end finetuned from the pretrained representation, 
		  shown in the top half of the table;
		`Frozen \cmark' means the pretrained representation is fixed and classified with a linear layer, `n.l.' denotes a non-linear classifier.
		For input, {\color{blue}`V' refers to visual only~(colored with blue)}, `A' is audio, `T' is text narration.
		MemDPC models with $\dagger$ refer to the two-stream networks, 
		where the predictions from RGB and Flow networks are averaged}
		}\label{table:sota}
		\begin{tabu}{ll|lccccc|cc} 
		\hline
		Method & Date & Dataset (duration) & Res. & Arch. & Depth &  Modality & Frozen & UCF  & HMDB  \\ \hline
		\rowfont{\color{blue}}
		CBT~\cite{Sun19}       & 2019\hspace{1pt}	& K600+ (273d)  & $112$     & S3D & 23   &V  & \cmark & 54.0 & 29.5 \\
		MIL-NCE~\cite{Miech20} & 2020\hspace{1pt}   & HTM (15y)     & $224$     & S3D & 23  &V+T    & \cmark & 82.7 & 53.1 \\
		MIL-NCE~\cite{Miech20} & 2020\hspace{1pt}   & HTM (15y)     & $224$     & I3D & 22  &V+T    & \cmark & 83.4 & 54.8 \\
		XDC~\cite{Alwassel19}  & 2019\hspace{1pt} 	& IG65M~(21y)   & $224 $  & R(2+1)D & 26  &V+A & \cmark & 85.3 & 56.0 \\ 
		ELO~\cite{Piergiovanni20}  & 2020\hspace{1pt} & Youtube8M- (8y)     & $224$   & R(2+1)D & 65    &V+A   & \cmark & -- & 64.5 \\ 
		\hline
		\textbf{MemDPC$\dagger$}& \hspace{1pt}   & K400 (28d)    & $224$     & R-2D3D & 33 &V    & \cmark & 54.1 & 30.5  \\ 
		\textbf{MemDPC$\dagger$}& \hspace{1pt}   & K400 (28d)    & $224$     & R-2D3D & 33 &V    & \cmark n.l. & 58.5 & 33.6  \\ 
		\hline \hline
	        \rowfont{\color{blue}}
		OPN~\cite{Lee17}        & 2017\hspace{1pt}       & UCF (1d)      & $227$      & VGG & 14   &V & \xmark & 59.6 & 23.8  \\ 
		\rowfont{\color{blue}}
		3D-RotNet~\cite{Jing18} & 2018\hspace{1pt}      & K400 (28d)    & $112$      & R3D & 17    &V & \xmark & 62.9 & 33.7  \\ 
		\rowfont{\color{blue}}
		ST-Puzzle~\cite{Kim19}  & 2019\hspace{1pt}       & K400 (28d)    & $224$      & R3D & 17    &V & \xmark & 63.9 & 33.7 \\ 
		\rowfont{\color{blue}}
		VCOP~\cite{Xu19vcop}    & 2019\hspace{1pt}       & UCF (1d)      & $112$     & R(2+1)D & 26 &V & \xmark & 72.4 & 30.9  \\ 
		\rowfont{\color{blue}}
		DPC~\cite{Han19}        & 2019 \hspace{1pt}     & K400 (28d)    & $224$     & R-2D3D & 33  &V & \xmark & 75.7 & 35.7  \\ 
		\rowfont{\color{blue}}
		CBT~\cite{Sun19}        & 2019\hspace{1pt}      & K600+ (273d)  & $112$     & S3D & 23     &V & \xmark & 79.5 & 44.6  \\
		\rowfont{\color{blue}}
		DynamoNet~\cite{Diba19} & 2019\hspace{1pt}    & Youtube8M-1 (1.9y)    & $112$     & STCNet & 133 &V & \xmark & 88.1 & 59.9  \\ 
		\rowfont{\color{blue}}
		SpeedNet~\cite{Benaim20} & 2020 \hspace{1pt}   & K400 (28d)    & $224$     & S3D-G & 23    &V & \xmark & 81.1 & 48.8  \\
		AVTS~\cite{Korbar18}    & 2018\hspace{1pt}    & K400 (28d)    & $224$     & I3D & 22     &V+A & \xmark & 83.7 & 53.0  \\
		AVTS~\cite{Korbar18}    & 2018 \hspace{1pt}   & AudioSet (240d)    & $224$     & MC3 & 17     &V+A & \xmark & 89.0 & 61.6  \\
		XDC~\cite{Alwassel19}   & 2019  \hspace{1pt}    & K400~(28d)   & $224 $     & R(2+1)D & 26   &V+A & \xmark & 84.2 & 47.1 \\ 
		XDC~\cite{Alwassel19}   & 2019  \hspace{1pt}    & IG65M~(21y)   & $224 $     & R(2+1)D & 26   &V+A & \xmark & 94.2 & 67.4 \\ 
		GDT~\cite{Patrick20}    & 2020 \hspace{1pt}     & K400 (28d)    & $112$     & R(2+1)D & 26   &V+A & \xmark & 88.7 & 57.8  \\ 
		MIL-NCE~\cite{Miech20}  & 2020\hspace{1pt}       & HTM (15y)     & $224$     & S3D-G  & 23  &V+T   & \xmark & 91.3 & 61.0  \\
		ELO~\cite{Piergiovanni20}  & 2020\hspace{1pt} & Youtube8M-2 (13y)     & $224$     & R(2+1)D & 65    &V+A   & \xmark & 93.8 & 67.4 \\ 
		\hline
		\textbf{MemDPC}         &  \hspace{1pt} & K400 (28d)    & $224$     & R-2D3D & 33 &V    & \xmark & 78.1 & 41.2   \\
		\textbf{MemDPC$\dagger$}& \hspace{1pt}  & K400 (28d)    & $224$     & R-2D3D & 33 &V    & \xmark &  86.1 & 54.5  \\
		\hline
		\rowfont{\color{mygray}}
		 Supervised ~\cite{Hara18} & \hspace{1pt}  & K400 (28d)    & $224$     & R3D & 33  &V    & \xmark & 87.7 & 59.1  \\
		\hline
		\end{tabu}
\end{table}

As shown in Table~\ref{table:sota}, 
for the same amount of data~(K400) 
and visual-only input, 
\MDPC\ surpasses all previous state-of-the-art self-supervised 
methods on both UCF101 and HMDB51
(although there exist small differences in architecture, 
\eg~for~3DRotNet, ST-Puzzle, DPC, SpeedNet).
When freezing the representation, 
it can be seen that a non-linear probe  gives better results than a linear probe, 
and in practice a non-linear classifier is still very cheap to train.

Other self-supervised training  methods on the same benchmarks are not directly comparable, 
even ignoring the architecture differences, due to the 
duration of videos used or  to the number of modalities used.
For example,
CBT~\cite{Sun19} uses a  longer version of K600 (referred to as K600+ in the table), the size is about 9 times that of the standard K400 that we use,
and CBT requires RotNet~\cite{Jing18} initialization while \MDPC\ can be trained from scratch. Nevertheless, 
our performance
exceeds that of CBT.
Other works use additional modalities for pre-text tasks like audio~\cite{Korbar18,Alwassel19,Patrick20,Piergiovanni20},
or narrations~\cite{Miech20}, and train on larger datasets.
Despite these disadvantages,
we demonstrate that
\MDPC\,  trained with only  visual inputs, can achieve competitive results on the finetuning protocol.

\subsection{Evaluation: Data Efficiency}
\label{exp:efficiency}
\begin{figure}[!htb]
	\centering
	\includegraphics[width=0.81\textwidth]{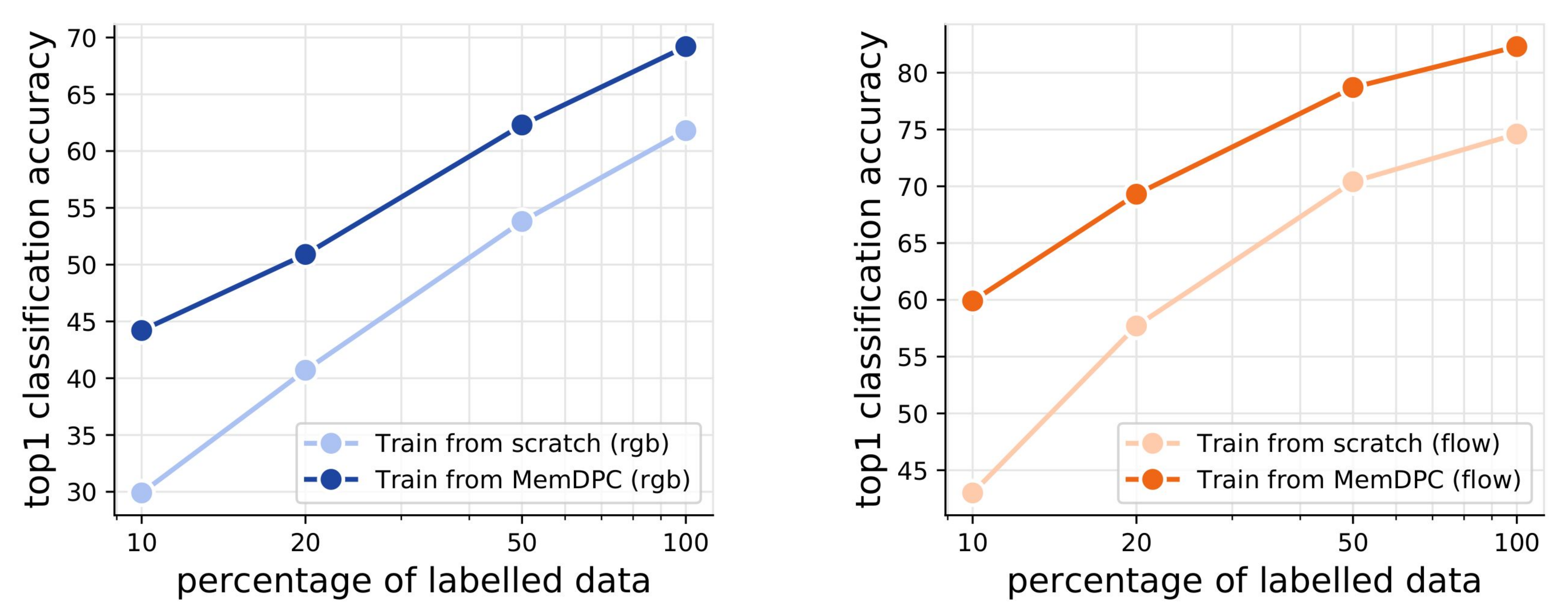}
	\caption{Data efficiency of \MDPC\ representations. 
	Left is RGB input and right is optical flow input. 
	The \MDPC\ is trained on UCF101 and it is evaluated on action classification~(finetuning protocol) on UCF101 with a reduced number of labels}\label{fig:data_efficiecy}
\end{figure}

In Figure~\ref{fig:data_efficiecy}, 
we show the data efficiency of \MDPC\ on both RGB input and optical flow with action recognition on the UCF101 dataset.
As we reduce the labelled training samples, 
action classifier trained on \MDPC\ representation generalize significantly better than the classifier trained from scratch.
Also, to match the performance of a random initialized classifier trained on 100\% labelled data, 
a classifier trained on \MDPC\ initialization only requires less than 50\% labelled data for both RGB and optical flow input. 

\subsection{Evaluation: Video Retrieval}
\label{exp:retrival}

In this protocol,  
we evaluate our representation with nearest-neighbour video retrieval,
features are extracted from the model, 
which is only trained with self-supervised learning, 
no further finetuning is allowed.

Experiments are shown on two datasets: UCF101 and HMDB51.
For both datasets, within the training set or within the testing set, 
multiple clips could be from the same source video,  hence they are visually similar and make the  retrieval task trivial. 
We follow the practice of~\cite{Xu19vcop,Luo20}, and
use each clip in the test set to query the $k$ nearest clips in the training set. 

For each clip, we sample multiple $8$ video blocks with a sliding window, 
and extract the context representation $c_t$ for each window. 
We spatial-pool each $c_t$ and take the average over all the windows. 
For distance measurement, we use cosine distance. 
We report Recall at $k$ (R@$k$) as the evaluation metric.
That is, as long as one clip of the same class is retrieved in the top $k$ nearest neighbours, 
a correct retrieval is counted.

\setlength{\tabcolsep}{4pt}
\begin{table}[h!]
\scriptsize
	\centering
	\caption{Comparison with others on Nearest-Neighbour video retrieval on UCF101 and HMDB51.
		Testing set clips are used to retrieve training set videos and R@$k$ is reported, where ${k}\in [1, 5, 10, 20]$. 
		Note that  all the models reported were only pretrained on UCF101 with self-supervised learning except SpeedNet}
		\begin{tabular}{lcc|cccc|cccc}
			\hline
			\multirow{2}{*}{Method} & \multirow{2}{*}{Date} & \multirow{2}{*}{Dataset} & \multicolumn{4}{c|}{UCF}         & \multicolumn{4}{c}{HMDB}       \\ \cline{4-11} 
			& & & R@1 & R@5 & R@10 & R@20  & R@1 & R@5 & R@10 & R@20  \\ \hline
			
	      Jigsaw~\cite{Noroozi16} & 2016 & UCF & 19.7 & 28.5 & 33.5  & 40.0    & -    & -    & -    & -    \\
		  OPN~\cite{Lee17} &     2017 & UCF    & 19.9 & 28.7 & 34.0  & 40.6  & -    & -    & -     & -      \\
		  Buchler~\cite{Buchler18}& 2018 & UCF& 25.7 & 36.2 & 42.2  & 49.2   & -    & -    & -     & -   \\
		 VCOP~\cite{Xu19vcop} & 2019 & UCF & 14.1 & 30.3 & 40.4  & 51.1    & 7.6  & 22.9 & 34.4  & 48.8    \\
		 VCP~\cite{Luo20} &  2020 & UCF     & 18.6 & 33.6 & 42.5  & 53.5   & 7.6  & 24.4 & 36.3  & 53.6   \\ 
		 SpeedNet~\cite{Benaim20} &  2020 & K400     & 13.0 & 28.1 & 37.5  & 49.5   & - & - & -  & -   \\ 
		\hline
		  \textbf{\MDPC}-RGB  & & UCF       & 20.2 & \textbf{40.4} & \textbf{52.4}  & \textbf{64.7}  & \textbf{7.7}  & \textbf{25.7} & \textbf{40.6}  & \textbf{57.7}   \\ 
		   \textbf{\MDPC}-Flow  & & UCF        & \textbf{40.2} & \textbf{63.2} & \textbf{71.9}  & \textbf{78.6}   & \textbf{15.6}  & \textbf{37.6} & \textbf{52.0}  & \textbf{65.3}  \\ 
		    \hline
		\end{tabular}
	\label{table:retrieval}
\end{table}


In Table~\ref{table:retrieval}, 
we show the retrieval performance on UCF101 and HMDB51. 
Note that the \MDPC\ benchmarked here is only trained on UCF101, the  same as~\cite{Xu19vcop,Luo20}. 
For fair comparison, \MDPC\ in this experiment uses a R18 backbone,
which has the same depth but  less parameters than the 3D-ResNet used in~\cite{Xu19vcop,Luo20}. 
With RGB inputs, 
our \MDPC\ gets state-of-the-art performance on all the metrics except R@1 in UCF101, 
where the method from Buchler~\etal~\cite{Buchler18} specializes well on R@1. 
While for Flow inputs, 
\MDPC{} significantly outperforms all previous methods by a large margin.
We also qualitatively show video retrieval results in the Appendix.

\subsection{Evaluation: Unintentional Actions}
\label{sec:oops}

We evaluate \MDPC\ on the Oops dataset on unintentional action classification.
In Oops, there is one failure moment in the middle of each video. 
When cutting the video into short clips, 
the clip overlapping the failure moment is defined as a `transitioning' action, 
the clips before are `intentional' actions, 
and the clips afterwards are `unintentional' actions. 
The core task is therefore to classify each short video clip into one of three categories,

In this experiment, 
we use a R18 based \MDPC\ model that
takes $128\times128$ resolution video frames as input.
After \MDPC\ is trained on K400 and the Oops training set videos with self-supervised learning, 
we further train it for unintentional action classification with a linear probe, 
and end-to-end finetuning~(as shown in Table~\ref{table:oops}).
The training details are given in the appendix.
State-of-the-art performance is demonstrated by our \MDPC\ on this unintentional action classification task,
even outperforming the model pretrained on K700 with full supervision with finetuning.

\begin{table}[!htb]
\footnotesize
	\centering 
	\caption{\MDPC\ on unintentional action classification tasks. 
	Note that our backbone 2+3D-ResNet18 has the same depth as 3D-ResNet18 used in~\cite{Epstein20} but with less parameters. 
	\MDPC\ model is trained on K400 and the OOPS training set without using labels, 
	and the network is then finetuned with supervision from the OOPS training set}
	\label{table:oops}
	\begin{tabular}{lll|ll}
		\hline
		Task               & Method                           & Backbone    & Freeze   & Finetune              \\ \hline
	    \multirow{3}{*}{Classification} & K700 Supervision                 & 3D-ResNet18 & \textbf{53.6} & 64.0            \\
		                   & Video Speed\cite{Epstein20}      & 3D-ResNet18 & 53.4 & 61.6            \\
		                   & \MDPC                            & R18         & 53.0    & \textbf{64.4}   \\ \hline
 
	\end{tabular}
\end{table}

\section{Conclusion}
In this paper, 
we propose a new architecture and learning framework~(\MDPC{}) for self-supervised learning from video, 
in particular for representations for action recognition.
With the novel compressive memory,
the model can efficiently handle the nature of multiple hypotheses in the self-supervised predictive learning procedure.
In order to thoroughly evaluate the quality of the learnt representation, 
we conduct experiments on four different downstream tasks, 
namely action recognition, video retrieval, learning with scarce annotations, and unintentional action classification.
In all cases,
we demonstrate state-of-the-art or competitive 
performance over other approaches that use orders of magnitude more training data.
Above all, 
for the first time, we show that it is possible to learn high-quality video representations with self-supervised learning, 
from the visual stream alone~(without additional audio or text streams).

\subsection*{Acknowledgements.}
Funding for this research is provided by a Google-DeepMind Graduate Scholarship, and by the EPSRC Programme Grant Seebibyte EP/M013774/1.
We would like to thank Jo\~{a}o F. Henriques, Samuel Albanie and Triantafyllos Afouras for helpful discussions. 

\clearpage
%
%
\bibliographystyle{splncs04}
\bibliography{bib/shortstrings,bib/vgg_local,bib/vgg_other}

\clearpage
\appendix
\noindent {\Large\textbf{Appendix}}
\section{Architectures in detail}

This section gives the architectural details of the \MDPC{} components, 
including the encoder $f(.)$ and temporal aggregator $g(.)$. 

\newcommand{\bred}[1]{\color{red}{\textbf{#1}}}
\paragraph{\textbf{Architecture of encoder $f(.)$.}}

The detailed architecture of the encoder function $f(.)$ is shown in Table~\ref{table:arch_f}. 
The size of the convolutional kernel is denoted by $[ \text{temporal}\times\text{spatial}^2 \text{, channel} ]$. 
The strides are denoted by $[ \text{temporal, }\text{spatial}^2 ]$.
We assume the input is 8 video blocks, 5 frames per video block, and the frame has a resolution of $128\times128$.
The column of `output size' shows the tensor dimension \textit{after} the current stage. 
Some pooling layers are omitted for clarity. We will release all the source code after the paper decision.

\begin{table}[!htb]
	\centering
	\caption{The 2D+3D ResNet18 structure of the encoding function $f(.)$. The 2D+3D ResNet34 structure replaces the depth of $\text{res}_2$ to $\text{res}_5$ from $\bred{2},\bred{2},\bred{2},\bred{2}$ to $\bred{3},\bred{4},\bred{6},\bred{3}$. }\label{table:arch_f}
	\begin{tabular}{c|c|c}
		\hline 
		stage   &  detail  & $ \begin{matrix} \text{output size}\\ T\times HW\times C \end{matrix}$ \\ \hline 
		input data & - & $5\times 128^2\times 3$ \\ \hline 
		$\text{conv}_1$  & $\begin{matrix} 1\times 7^{2}, 64\\ \text{stride } 1, 2^2\end{matrix}$ & $5\times64^2\times64$ \\ \hline 
		$\text{pool}_1$    & $\begin{matrix} 1\times 3^{2}, 64\\ \text{stride } 1, 2^2\end{matrix}$ & $5\times32^2\times64$ \\ \hline  
		$\text{res}_2$     & $\begin{bmatrix} 1\times3^2, 64\\1\times3^2, 64 \end{bmatrix}\times \bred{2}$ & $5\times32^2\times64$ \\ \hline  
		$\text{res}_3$     & $\begin{bmatrix} 1\times3^2, 128\\1\times3^2, 128 \end{bmatrix}\times \bred{2}$ & $5\times16^2\times128$ \\ \hline  
		$\text{res}_4$     & $\begin{bmatrix} 3\times3^2, 256\\3\times3^2, 256 \end{bmatrix}\times \bred{2}$ & $3\times8^2\times256$ \\ \hline  
		$\text{res}_5$     & $\begin{bmatrix} 3\times3^2, 256\\3\times3^2, 256 \end{bmatrix}\times \bred{2}$ & $2\times4^2\times256$ \\ \hline  
		$\text{pool}_2$    & $\begin{matrix} 2\times 1^{2}, 256\\ \text{stride } 1, 1^2\end{matrix}$ & $1\times4^2\times256$ \\ 
		\hline 
	\end{tabular}
	
\end{table}


\paragraph{\textbf{Architecture of temporal aggregator $g(.)$.}}

The detailed architecture of the temporal aggregation function $g(.)$ is shown in Table~\ref{table:arch_g}. It aggregates the feature maps over the past $T$ time steps. 
Table~\ref{table:arch_g} shows the case where $g(.)$ aggregates the feature maps over the past 5 steps. 
We use the same convention as above to denote convolutional kernel size.
The temporal aggregator is an one-layer ConvGRU that returns context feature with same number of channels as its input. 

\begin{table}[t]
	\centering
		\caption{The structure of aggregation function $g(.)$. }\label{table:arch_g}
	\begin{tabular}{c|c|c}
		\hline 
		stage  & detail & $\begin{matrix}\text{output sizes}\\T\times t\times d^2\times C\end{matrix}$ \\ \hline 
		input data & - & $5\times1\times4^2\times256$\\ \hline 
		ConvGRU &$\begin{matrix} [1^2, 256]\end{matrix}\times1 \text{ layer}$ & $1\times1\times4^2\times256$ \\
		\hline  
	\end{tabular}
\end{table}

\section{Details of unintentional action classification}

\begin{figure}[h!]
	\centering
	\includegraphics[width=0.8\textwidth]{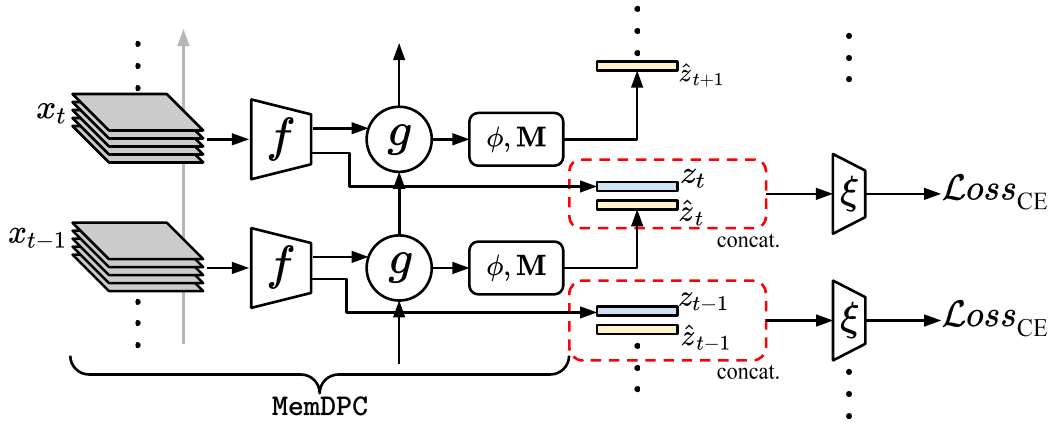}
	\caption{Architecture of the unintentional action classification framework.}\label{fig:oops_arch}
\end{figure}

Figure~\ref{fig:oops_arch} shows the architecture of the unintentional action classification task. 
In detail, at each time step $t$, the single-directional \MDPC{} framework produces a feature $z_{t}$ from the current input $x_t$,
and also a predicted feature $\hat{z}_{t}$ from the past input $x_1, ..., x_{t-1}$.
For each time step, two features $z_t$ and $\hat{z}_{t}$ are concatenated and passed to a linear classifier $\xi(.)$
to classify into one of three categories (intentional, transitioning or unintentional actions). 

Naturally in the Oops dataset the distribution of the three classes is very unbalanced, \eg{} transitioning actions are very rare comparing with the other two categories.
To handle this we oversample transitioning actions during training.
For testing, we take the same sequence length as training from the video with a temporal moving window, and summarize the prediction. 

The classifier is trained with cross entropy loss and optimized by the Adam optimizer with a $10^{-3}$ learning rate. 
The learning rate is decayed once to $10^{-4}$ when the validation loss plateaus. 
\clearpage

\section{Video retrieval results}

\begin{figure}[!htb]
	\centering
	\includegraphics[width=.9\textwidth]{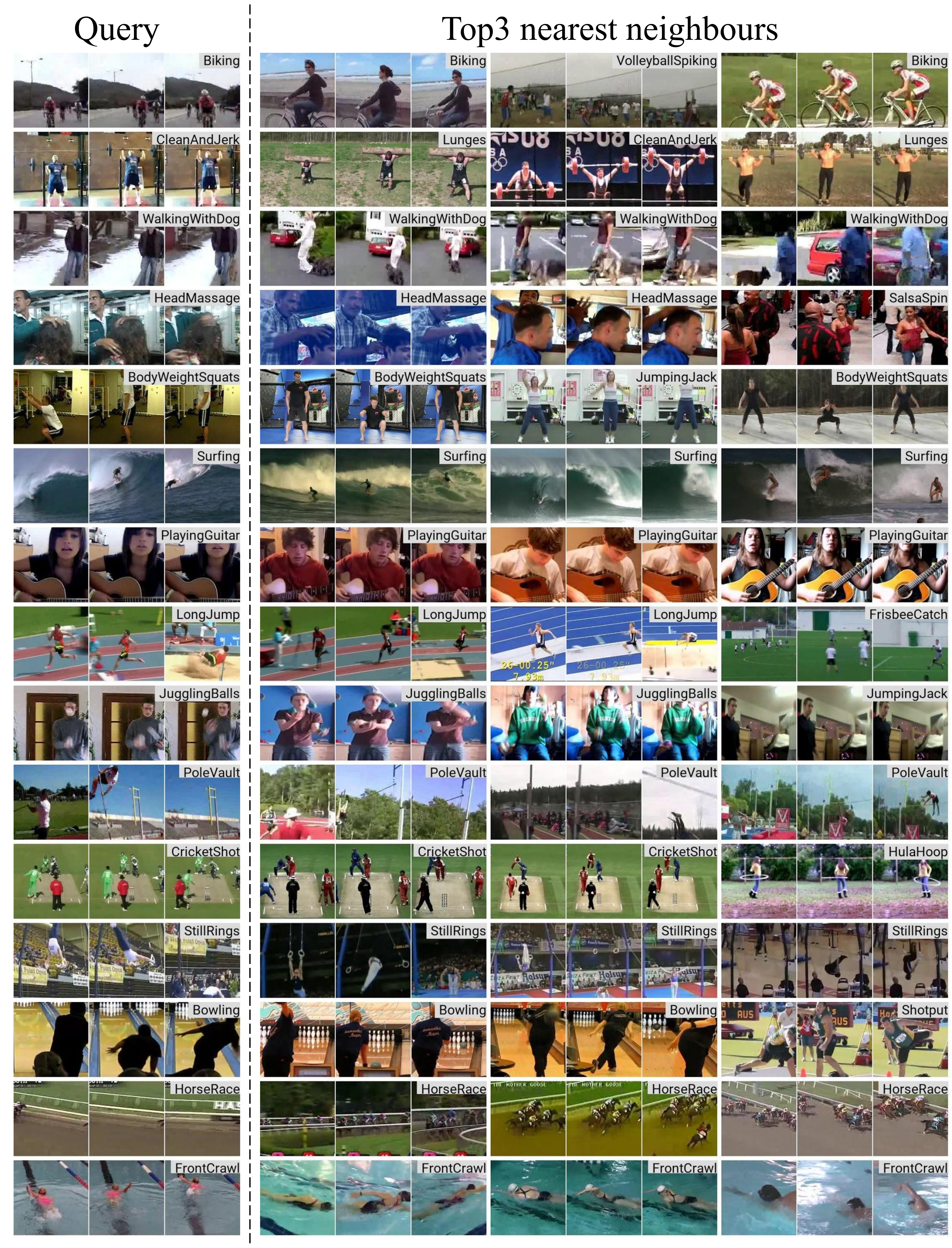}
	\caption{More nearest neighbour retrieval results with \MDPC\ representations. The left side is the query video from the UCF101 testing set, and the right side are the top 3 nearest neighbours from the UCF101 training set. The \MDPC\ is trained only on UCF101 with optical flow input and we visualize their raw RGB frames for clarity.
	The action label for each video is shown in the upper right corner.}\label{fig:more_retrieval}
\end{figure}

\section{Pseudocode of MemDPC}

In this section, we give the core implementation of \MDPC{} in PyTorch-like style, including the compressive memory bank and the computation of the contrastive loss. We will release all the source code. 

\newcommand\algcomment[1]{\def\@algcomment{\footnotesize#1}}

\begin{algorithm}[h]
\caption{Pseudocode for \MDPC{} in PyTorch-like style.}\label{code}
\begin{lstlisting}[language=python]
# f: feature extractor, 2d3d-ResNet
# g: aggregator, ConvGRU
# phi: future prediction function, MLP
# x: video input, size = [BatchSize,NumBlock,BlockSize,Channel,Height,Width], 
                    # e.g. [16,8,5,3,128,128]
# pred_step: prediction step, e.g. predict 3 steps into the future
# MB: the compressive memory bank, has size [k,C], e.g. [1024, 256]

z_hat_all = []
feature_z = f(x) # extract feature,
                   # size=[B,N,C,H,W], e.g. [16,8,256,4,4]

for i in range(pred_step): # sequentially predict into the future
    if i == 0: 
        feature_z_tmp = feature_z[:,0:(N-pred_step),:] # get past features
        hidden = g(feature_z_tmp) # temporal aggregation, hidden state == context, 
                                     # size=[B,C,H,W], e.g. [16,256,4,4]
    else:
        hidden = g(feature_z_hat, hidden) # aggregate one more step,
                                              # size=[B,C,H,W]

    prob = Softmax(phi(hidden), dim=1)  # probability over MB, 
                                           # size=[B,k,H,W], e.g. [16,1024,4,4]
    feature_z_hat = torch.einsum('bkhw,kc->bchw', prob, MB) # weighted sum over MB, 
                                                                  # size=[B,C,H,W]
    z_hat_all.append(feature_z_hat)

z_hat = torch.stack(z_hat_all, dim=1) # predicted feature, size=[B,pred_step,C,H,W], 
                                          # e.g. [16,3,256,4,4]
z = feature_z[:,-pred_step::,:]       # desired feature, size=[B,pred_step,C,H,W]

# dot product over 'C', and flatten as 2D matrix, size=[B*pred_step*H*W, B*pred_step*H*W]
similarity = torch.einsum('abcde,fgchi->abdefghi', z_hat, z)\
                    .view(B*pred_step*H*W, B*pred_step*H*W) 
target = torch.arange(B*pred_step*H*W) # diagonal of similarity matrix is positive

loss = CrossEntropyLoss(similarity, target)
loss.backward()
\end{lstlisting}
\end{algorithm}

\clearpage

\section{Visualization of learned memory}

\vspace{-20pt}
\begin{figure}[!htb]
	\centering
	\includegraphics[width=0.95\textwidth]{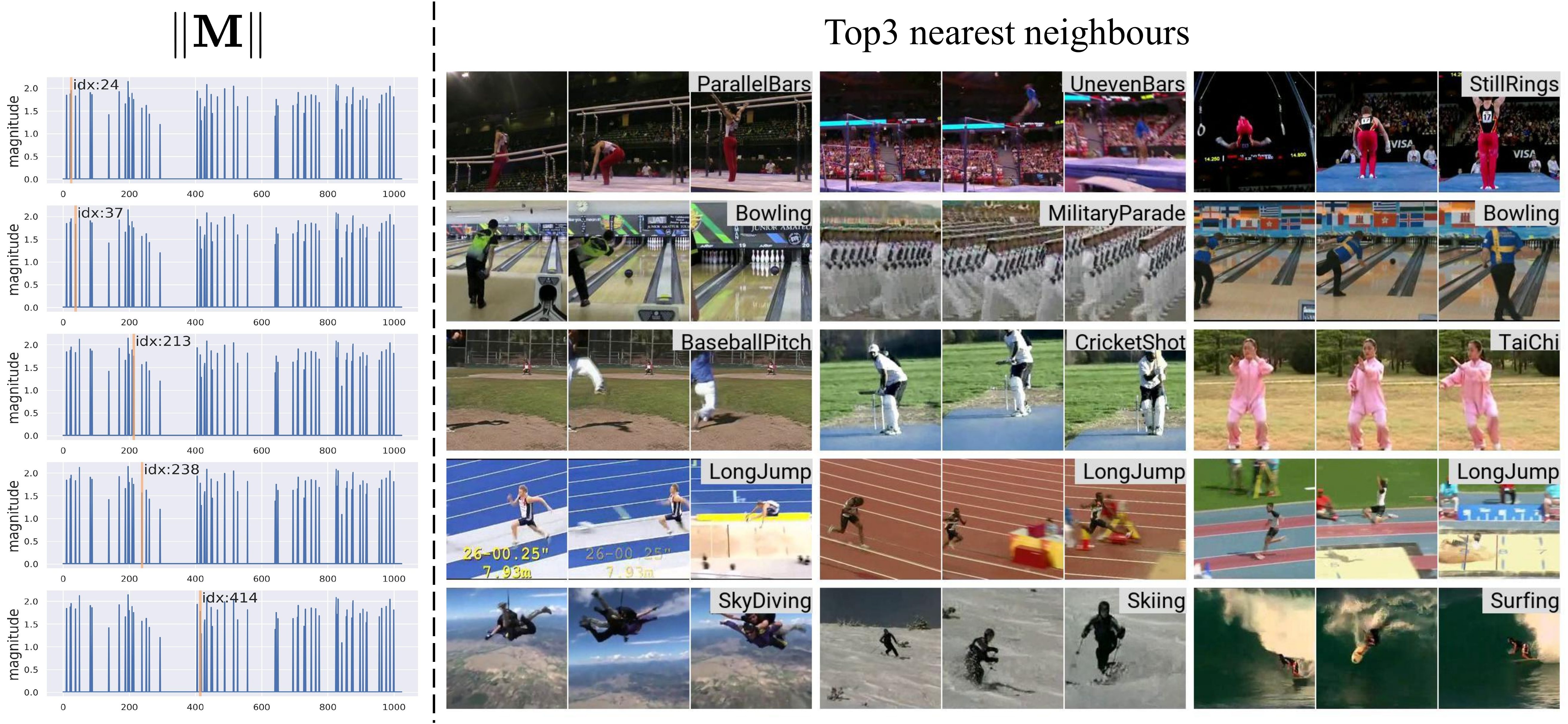}
	\caption{Nearest-neighbour retrieval for $\{24, 37, 213, 238, 414\}$-th memory entries with UCF101 training videos. The left shows the magnitude of memory bank, $\| \mathbf{M} \| \in \mathbb{R}^{1024}$, with the chosen memory entry highlighted in orange. The right shows the top-3 nearest neighbour in the UCF101 training set retrieved by the corresponding memory entry. }\label{fig:mem-retrieval}
\end{figure}

\vspace{-20pt}
\begin{figure}[h!]
	\centering
	\includegraphics[width=0.95\textwidth]{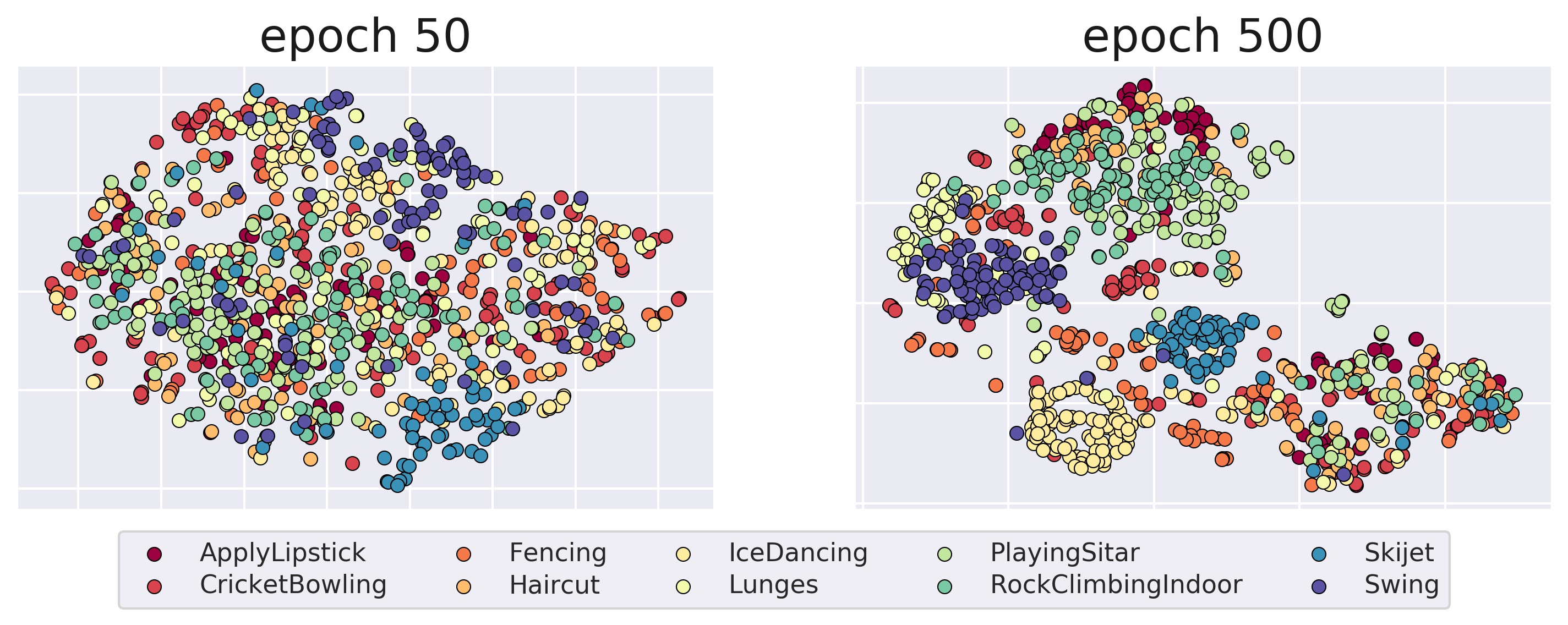}
	\caption{t-SNE visualization of $p_{t+1}$ from paper Eq.~\ref{eq:pmam} at different training stages. For clarity, 10 action classes are randomly chosen from UCF101 training set and visualized.}\label{fig:mem-tsne}
\end{figure}

This section visualizes the memory learned by \MDPC. In Figure~\ref{fig:mem-retrieval}, we use single memory entry as the query to retrieve videos in the feature space. It shows the memory may have captured certain features in the video like repetitive textures or wide background. Figure~\ref{fig:mem-tsne} shows the t-SNE clustering results of the memory addressing probability $p_{t+1}$ from Equation~\ref{eq:pmam}. It shows that as the training progresses, the network attends to different memory entries to predict futures for different action categories, although category information is \emph{not} involved during training.

\end{document}